\documentclass[10pt,twocolumn,letterpaper]{article}

\usepackage{iccv}
\usepackage{times}
\usepackage{epsfig}
\usepackage{graphicx}
\usepackage{amsmath}
\usepackage{amssymb}
\usepackage[export]{adjustbox}

\def\etal{{\em et al.}}
\usepackage[font=small,skip=5pt]{caption}
\usepackage{float}

\usepackage{multirow}
\usepackage{soul}
\usepackage{subfigure}
\usepackage{array}
\usepackage{comment}

\usepackage[mathscr]{euscript}
\newcolumntype{L}[1]{>{\raggedright\let\newline\\arraybackslash\hspace{0pt}}m{#1}}
\newcolumntype{C}[1]{>{\centering\let\newline\\arraybackslash\hspace{0pt}}m{#1}}
\newcolumntype{R}[1]{>{\raggedleft\let\newline\\arraybackslash\hspace{0pt}}m{#1}}
\usepackage{booktabs}
\usepackage{enumitem}

\usepackage[pagebackref=true,breaklinks=true,letterpaper=true,bookmarks=false]{hyperref}
\renewcommand{\vec}[1]{\boldsymbol{#1}}
\newcommand{\mat}[1]{\mathbf{#1}}
\newcommand{\set}[1]{\mathcal{#1}}

\newcommand{\imageset}[0]{\set{I}}
\newcommand{\image}[0]{\mat{I}}

\newcommand{\poseset}[0]{\set{P}}

\newcommand{\jointset}[0]{\set{J}}
\newcommand{\garmparamset}[0]{\set{G}}

\newcommand{\template}[0]{\mat{T}}

\newcommand{\blendweights}[0]{\mat{W}}

\newcommand{\pose}[0]{\vec{\theta}}
\newcommand{\shape}[0]{\vec{\beta}}
\newcommand{\trans}[0]{\vec{t}}
\newcommand{\joints}[0]{\mat{J}}
\newcommand{\garmparam}[0]{\vec{z}}
\newcommand{\offsets}[0]{\mathbf{D}}

\newcommand{\smpl}[0]{M}
\newcommand{\posefun}[0]{T}
\newcommand{\blendfun}[0]{W}
\newcommand{\offsetfun}[0]{B}
\newcommand{\jointfun}[0]{J}

\newcommand{\basenet}[0]{f_w}
\newcommand{\basenetstar}[0]{f_w^{*}}

\newcommand{\garmnet}[0]{M_w^g}

\newcommand{\loss}[0]{\mathcal{L}}
\newcommand{\myparagraph}[1]{\noindent\vspace{1pt}{\textbf{#1}}}
\graphicspath{ {./images/} }

\iccvfinalcopy %

\def\iccvPaperID{4019} %

\ificcvfinal\fi

\begin{document}
\title{Multi-Garment Net: Learning to Dress 3D People from Images}

\author{Bharat Lal Bhatnagar\hspace{8mm}
	 Garvita Tiwari\hspace{8mm}
	Christian Theobalt\hspace{8mm}
	Gerard Pons-Moll \\
Max Planck Institute for Informatics, Saarland Informatics Campus, Germany\\
{\tt\small \{bbhatnag,gtiwari,theobalt,gpons\}@mpi-inf.mpg.de}
}

\makeatletter

\makeatother

\maketitle
\pagestyle{plain}

\begin{abstract}
    We present Multi-Garment Network (MGN), a method to predict body shape and clothing, layered on top of the SMPL~\cite{smpl2015loper} model from a few frames (1-8) of a video. 
Several experiments demonstrate that this representation allows higher level of control when compared to single mesh or voxel representations of shape. 
Our model allows to predict garment geometry, relate it to the body shape, and transfer it to new body shapes and poses.
To train MGN, we leverage a digital wardrobe containing 712 digital garments in correspondence, obtained with a novel method to register a set of clothing templates to a dataset of real 3D scans of people in different clothing and poses. Garments from the digital wardrobe, or predicted by MGN, can be used to dress any body shape in arbitrary poses. 
We will make publicly available the digital wardrobe, the MGN model, and code to \emph{dress} SMPL with the garments at \cite{MGNcode}.
\end{abstract}

\section{Introduction}
The 3D reconstruction and modelling of humans from images is a central problem in computer vision and graphics.
Although a few recent methods~\cite{alldieck2018video,alldieck2019learning,alldieck2018detailed,habermann2019TOG,natsume2018siclope,saito2019pifu} attempt reconstruction of people with clothing, they lack realism and control. This limitation is in great part due to the fact that they use a single surface (mesh or voxels) to represent both clothing and body. Hence they can not capture the clothing separately from the subject in the image, let alone map it to a novel body shape.

\begin{figure}[t]
	\centering
	\includegraphics[width=0.48\textwidth]{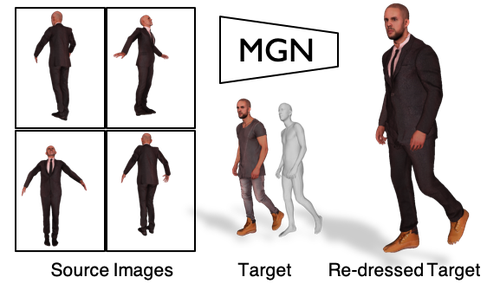}
	\caption{Garment re-targeting with Multi-Garment Network (MGN). Left to right: images from source subject, body from the target subject, target dressed with source garments. From one or more images, MGN can reconstruct the body shape and each of the garments separately. We can transfer the predicted garments to a novel body including geometry and texture.}
	\label{fig:teaser}
\end{figure}

In this paper, we introduce Multi-Garment Network (MGN), the first model capable of inferring human body and layered garments on top as separate meshes from images directly.
As illustrated in Fig.~\ref{fig:teaser} this new representation allows full control over body shape, texture and geometry of clothing and opens the door to a range of applications in VR/AR, entertainment, cinematography and virtual try-on.

Compared to previous work, MGN produces reconstructions of higher visual quality, and allows for more control: 1) we can infer the 3D clothing from one subject, and dress a second subject with it, (see Fig.~\ref{fig:teaser}, \ref{fig:MGN_retarget}) and 2) we can trivially map the garment texture captured from images to any garment geometry of the same category (see Fig.\ref{fig:tex-transfer}).

To achieve such level of control, we address two major challenges: learning per-garment models from 3D scans of people in clothing, and learning to reconstruct them from images.  
We define a discrete set of garment templates (according to the categories long/short shirt, long/short pants and coat) and register, for every category, a single template to each of the scan instances, which we automatically segmented into clothing parts and skin. 
Since garment geometry varies significantly within one category (\emph{e.g.} different shapes, sleeve lengths), we first minimize the distance between template and the scan boundaries, while trying to preserve the Laplacian of the template surface. This initialization step only requires solving a linear system, and 
nicely stretches and compresses the template globally, which we found crucial to make subsequent non-rigid registration work. 
Using this, we compile a \emph{digital wardrobe} of real 3D garments worn by people, (see Fig.~\ref{fig:Digital wardrobe}). From such registrations, we learn a vertex based PCA model per garment. Since garments are naturally associated with the underlying SMPL body model, we can transfer them to different body shapes, and re-pose them using SMPL. 
\begin{figure*}[t]
	\centering
	\includegraphics[width=1\linewidth]{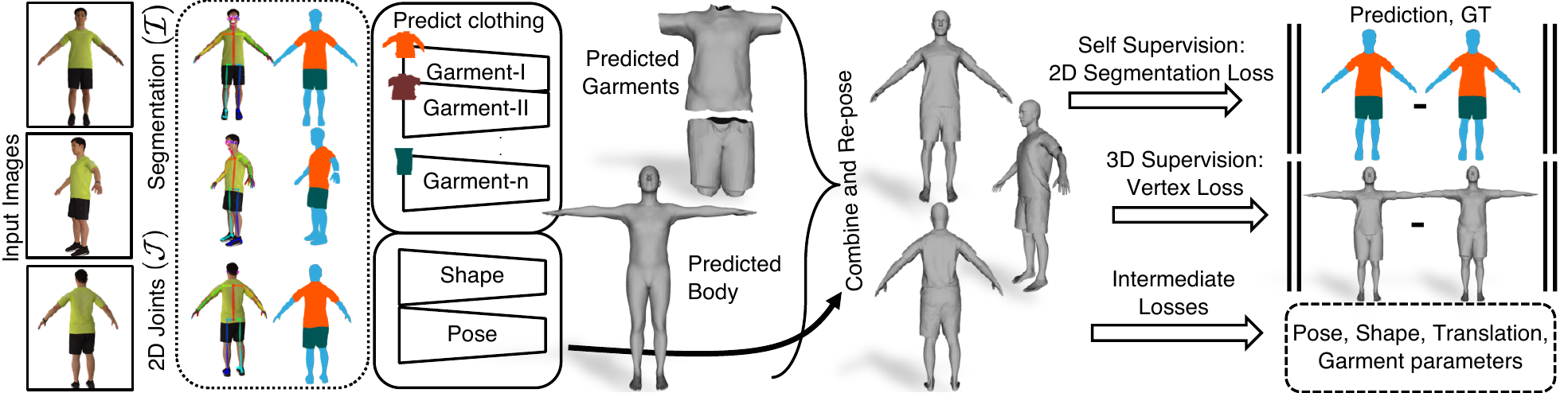}
	\caption{ Overview of our approach.
	Given a small number of RGB frames (currently 8), we pre-compute semantically segmented images ($\imageset$) and $2D$ Joints ($\jointset$). Our Multi-Garment Network (MGN), takes $\{\imageset,  \jointset\}$ as input and infers separable garments and the underlying human shape in a canonical pose. We repose these predictions using our per-frame pose predictions. We train MGN with a combination of 2D and 3D supervision. The 2D supervision can be used for online refinement at test time.}
	\label{fig:overview}
\end{figure*}
From the digital wardrobe, MGN is trained to predict, given one or more images of the person, the body pose and shape parameters, the PCA coefficients of each of the garments, and a displacement field on top of PCA that encodes clothing detail. At test time, we refine this bottom-up estimates with a new top-down objective that forces projected garments and skin to explain the input semantic segmentation. This allows more fine-grained image matching as compared to standard silhouette matching.
Our contributions can be summarized as: 
\begin{itemize}[leftmargin=*]
	\item A novel data driven method to infer, for the first time, separate body shape and clothing from just images (few RGB images of a person rotating in front of the camera).
	\vspace{-0.12cm}
	\item A robust pipeline for 3D scan segmentation and registration of garments. To the best of our knowledge, there are no existing works capable of automatically registering a single garment template set to multiple scans of real people with clothing.
	\vspace{-0.12cm}
	\item A novel top-down objective function that forces the predicted garments and body to fit the input semantic segmentation images. 
	\vspace{-0.12cm}
	\item We demonstrate several applications that were not previously possible such as dressing avatars with predicted $3D$ garments from images, and transfer of garment texture and geometry.
	\vspace{-0.12cm}
	\item We will make publicly available the MGN to predict $3D$ clothing from images, the digital wardrobe, as well as code to ``dress'' SMPL with it. 
\end{itemize}

\section{Related Work}
In this section we discuss the two branches of work most related to our method, namely \emph{capture} of clothing and body shape and \emph{data-driven} clothing models.

\myparagraph{Performance Capture.}
The classical approach to bring dynamic sequences into correspondence is to deform meshes non-rigidly~\cite{carranza2003free,deAguiar2008performance,cagniart_meshdeform} or volumetric shape representations~\cite{huang2016volumetric,InriaVolumetric_2015} to fit multiple image silhouettes. 
Without a pre-scanned template, fusion~\cite{izadi2011kinectfusion,innmann2016volume,slavcheva2017killingfusion,newcombe2015dynamicfusion,DoubleFusion2018} trackers incrementally fuse geometry and appearance~\cite{zhou2014color} to build the template on the fly. Although flexible, these require multi-view~\cite{starck2007surface,inria_2017,collet2015high}, one or more depth cameras~\cite{dou2016fusion4d,orts2016holoportation}, or require the subject to stand still while turning the cameras around them~\cite{shapiro2014rapid,3Dportraits,zeng2013templateless,cui2012kinectavatar}. From RGB video, Habermann \emph{et al.}\cite{habermann2019TOG} introduced a real time tracking system to capture non-rigid clothing dynamics.
Very recently,
SimulCap~\cite{SimulCap19} allows multi-part tracking of human performances from a depth camera.

\myparagraph{Body and cloth capture from images and depth.} 
Since current statistical models can not represent clothing, most works~\cite{anguelov2005scape,hasler2009statistical,smpl2015loper,zuffi2015stitched,pons2015dyna,joo2018total,guan2009estimating,zhou2010parametric,jain2010moviereshape,rogge2014garment,bogo2016smplify, kanazawa2018endtoend,omran2018neural, pavlakos2018humanshape} are restricted to inferring body shape alone.
Model fits have been used to virtually dress and manipulate people's shape and clothing in images~\cite{rogge2014garment,zhou2010parametric,zanfir2018human,Lassner:GP:2017}. None of these approaches recover 3D clothing. Estimating body shape and clothing from an image has been attempted in~\cite{guo2012clothed,chen2013deformable}, but it does not separate clothing from body and requires manual intervention ~\cite{zhou2013garment, robson2011context} .
Given a depth camera, Chen~\emph{et al.} \cite{chen2015garment} retrieve similar looking synthetic clothing templates from a database. Dan{\v{e}}{\v{r}}ek \emph{et al.} \cite{danvevrek2017deepgarment} use physics based simulation to train a CNN but do not estimate garment and body jointly, require pre-specified garment type, and the results can only be as good as the synthetic data.

Closer to ours is the work of Alldieck \emph{et al.}~\cite{alldieck2019learning,alldieck2018video,alldieck2019tex2shape} which reconstructs, from a single image or a video, clothing and hair as displacements on top of SMPL, but can not separate garments from body, and can not transfer clothing to new subjects.
In stark contrast to ~\cite{alldieck2019learning}, we register the scan garments (matching boundaries) and body separately, which allows us to learn the mapping from images to a multi-layer representation of people. 

\myparagraph{Data-driven clothing.} 
A common strategy to learn efficient data-driven models is to use off-line simulations ~\cite{deAguiar:2010:SSR,Kim:2013:NEP,DRAPE2012,sigal2015perceptual,Santesteban2019,gundogdu2018garnet} for generating data.
These approaches often lack realism when compared to models trained using real data.
Very few approaches have shown models learned from real data. Given a dynamic scan sequence, Neophytou \etal \cite{Neophytou2014layered} learn a two layer model (body and clothing) and use it to dress novel shapes.  A similar model has been recently proposed~\cite{yang2018analyzing}, where the clothing layer is associated to the body in a fuzzy fashion. Other methods~\cite{yang2016estimation,shapeundercloth:CVPR17} focus explicitly on estimating the body shape under clothing.  Like these methods, we treat the underlying body shape as a layer, but unlike them, we segment out the different garments allowing sharp boundaries and more control.
For garment registration, we build on the ideas of ClothCap~\cite{ponsmoll2017clothcap}, which can register a \emph{subject specific} multi-part model to a 4D scan sequence. By contrast, we register a single template set to multiple scan instances-- varying in garment geometry, subject identity and pose, which requires a new solution. 
Most importantly, unlike all previous work~\cite{lahner2018deepwrinkles,ponsmoll2017clothcap,yang2018analyzing}, we learn per-garment models and train a CNN to predict body shape and garment geometry directly from images.

\section{Method}
\begin{figure*}[t]
	\centering
	\includegraphics[width=\textwidth]{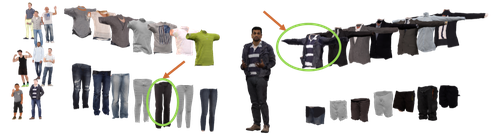}
	\caption{Digital 3D wardrobe. We use our proposed multi-mesh registration approach to register garments present in the scans (left) to fixed garment templates. This allows us to build a digital wardrobe and dress arbitrary subjects (center) by picking the garments (marked) from the wardrobe.}
	\label{fig:Digital wardrobe}
\end{figure*}
In order to learn a model to predict body shape and garment geometry directly from images, we process a dataset of $356$ scans of people in varied clothing, poses and shapes. Our data pre-processing (Sec.~\ref{sec:pre-processing}) consists of the following steps: SMPL registration to the scans, body aware scan segmentation and template registration. 
We obtain, for every scan, the underlying body shape, and the garments of the person registered to one of the $5$ garment template categories: shirt, t-shirt, coat, short-pants, long-pants. The obtained digital wardrobe is illustrated in Fig.~\ref{fig:Digital wardrobe}.  The garment templates are defined as regions on the SMPL surface; the original shape follows a human body, but it deforms to fit each of the scan instances after registration. Since garment registrations are naturally associated to the body represented with SMPL, they can be easily reposed to arbitrary poses. With this data, we train our Multi-Garment Network to estimate the body shape and garments from one or more images of a person, see Sec.~\ref{sec:MGN}.

\subsection{Data Pre-Processing: Scan Segmentation and Registration}
\label{sec:pre-processing}
Unlike ClothCap~\cite{ponsmoll2017clothcap} which registers a template to a $4D$ scan sequence of a \emph{single subject}, our task is to register single template across instances of varying styles, geometries, body shapes and poses. Since our registration follows the ideas of ~\cite{ponsmoll2017clothcap}, we describe the main differences here.

\myparagraph{Body-Aware Scan Segmentation} 
\begin{figure}[t]
	\centering
	\includegraphics[width=0.4\textwidth]{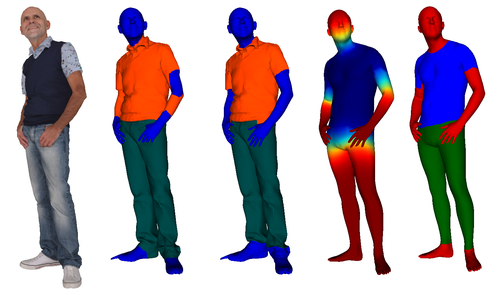}
	\caption{Left to right: Scan, segmentation with MRF and CNN unaries, MRF with CNN unaries + garment prior + appearance terms, the garment(t-shirt) prior based on geodesics and the template. Notice how the garment prior is crucial to obtain robust results.}
	\label{fig:Segmentation}
\end{figure}
We first automatically segment the scans into three regions: skin, upper-clothes and pants (we annotate the garments present for every scan). 
Since even SOTA image semantic segmentation~\cite{gong2018instance} is inaccurate, naive lifting to 3D is not sufficient. Hence, we incorporate \emph{body specific garment priors} and segment scans by solving an MRF on the UV-map of the SMPL surface after non-rigid alignment. 

A garment prior (for garment $g$) derives from a set of labels $\mathbf{l}^i_g \in \{ 0,1 \}$ indicating the vertices $\mathbf{v}_i \in \mathcal{S}$ of SMPL that are likely to overlap with the garment. 
The aim is to penalize labelling vertices as $g$ outside this region, see Fig~\ref{fig:Segmentation}. Since garment geometry varies significantly within one category (\eg t-shirts of different sleeve lengths), we define a cost increasing with the geodesic distance $\mathrm{dist}_{\mathrm{geo}}(\mathbf{v}): \mathcal{S} \mapsto \mathbb{R}$ from the garment region boundary -- efficiently computed based on heat flow~\cite{crane2013geodesics}. 
Conversely, we define a similar penalty for labeling vertices in the garment region with a label different than $g$. As data terms, we incorporate CNN based semantic segmentation~\cite{gong2018instance}, and appearance terms based Gaussian Mixture Models in \emph{La} color space. The influence of each term is illustrated in Fig.~\ref{fig:Segmentation}, for more details we refer to the supp. mat.

After solving the MRF on the SMPL UV map, we can segment the scans into 3 parts by transferring the labels from the SMPL registration to the scan.

\myparagraph{Garment Template}
We build our garment template on top of SMPL+D, $\smpl(\cdot)$, which represents the human body as a parametric function of pose($\pose$), shape($\shape$), global translation($\trans$) and optional per-vertex displacements ($\offsets$):
\begin{equation}
\label{eq:smplpose}
\smpl(\shape,\pose,\offsets) = \blendfun(\posefun(\shape,\pose,\offsets), \jointfun(\shape), \pose, \blendweights)
\end{equation}
\begin{equation}
\label{eq:smplshape}
\posefun(\shape,\pose,\offsets) = \template + \offsetfun_s(\shape) + \offsetfun_p(\pose) + \offsets.
\end{equation}
The basic principle of SMPL is to apply a series of linear displacements to a base mesh $\template$  with $n$ vertices in a T-pose, and then apply standard skinning $W(\cdot)$. 
Specifically, $\offsetfun_p(\cdot)$ models pose-dependent deformations of a skeleton $\jointfun$, and $\offsetfun_s(\cdot)$ models the shape dependent deformations. $\blendweights$ represents the blend weights.

For each garment class $g$ we define a template mesh, $\mat{G}^g$ in T-pose, which we subsequently register to explain the scan garments.
We define $\mat{I}^g \in \mathbb{Z}^{m_g\times{n}}$ as an indicator matrix, with $\mathbf{I}^g_{i,j}=1$ if garment $g$ vertex $i \in \{1 \hdots m_g\}$ is associated with body shape vertex $j \in \{1 \hdots n\}$. In our experiments, we associate a single body shape vertex to each garment vertex.
We compute displacements to the corresponding SMPL body shape $\shape^g$ under the garment as 
\begin{equation}
\label{eq:disp}
\mathbf{D}^g = \mat{G}^g - \mathbf{I}^g \posefun(\shape^g,\mat{0_\theta},\mat{0_D})
\end{equation}
Consequently, we can obtain the garment shape (unposed), $T^g$ for a new shape $\shape$ and pose $\pose$ as
\begin{equation}
\label{eq:garment_shape}
\posefun^g(\shape,\pose,\mathbf{D}^g) = \mat{I}^g\posefun(\shape,\pose,\mat{0})+\mathbf{D}^g
\end{equation}
To pose the vertices of a garment, each vertex uses the skinning function in Eq.~\ref{eq:smplpose} of the associated SMPL body vertex.
\begin{equation}
\label{eq:garment_pose}
    G(\shape,\pose,\mathbf{D}^g) = \blendfun(\posefun^g(\shape,\pose,\offsets^g), \jointfun(\shape), \pose, \blendweights)
\end{equation}

\myparagraph{Garment Registration}
Given the segmented scans, we non-rigidly register the body and garment templates (upper-clothes, lower-clothes) to scans using the multi-part alignment proposed in~\cite{ponsmoll2017clothcap}. The challenging part is that garment geometries vary significantly across instances, which makes the multi-part registration fail (see supplementary). 
Hence, we first initialize by deforming the vertices of each garment template with the shape and pose of SMPL registrations, obtaining deformed vertices $\mat{G}_{\mathrm{init}}^g$.
Note that since the vertices defining each garment template are fixed, the clothing boundaries of the initially deformed garment template will not match the scan boundaries. In order to globally deform the template to match the clothing boundaries in a single shot, we define an objective function based on Laplacian deformation~\cite{sorkine2005laplacian}.

Let $\mat{L}^g\in \mathbb{R}^{m_g\times{m_g}}$ be the graph Laplacian of the garment mesh, and $\boldsymbol{\Delta}_{\mathrm{init}} \in\mathbb{R}^{m_g\times{3}}$ the differential coordinates of the initially deformed garment template $\boldsymbol{\Delta}_{\mathrm{init}} = \mathbf{L}\,\mat{G}_{\mathrm{init}}^g$. For every vertex $\mathbf{s}_i \in \mathcal{S}_b$ in a scan boundary $\mathcal{S}_b$, we find its closest vertex in the corresponding template garment boundary, obtaining a matrix of scan points $\mathbf{q}_{1:C} = \{\mathbf{q}_1, \hdots,\mathbf{q}_C\}$ with corresponding template vertex indices $j_{1:C}$. Let $\mathbf{I}_{C\times{m_g}}$ be a selector matrix indicating the indices in the template corresponding to each $\mathbf{q}_i$. With this, we minimize the following least squares problem: 
\begin{equation}
\begin{bmatrix}
	\mathbf{L}^g
	\\ 
	w\mathbf{I}_{C\times{m_g}}
\end{bmatrix} \mathbf{G}^g = 
\begin{bmatrix}
	\boldsymbol{\Delta}_{\mathrm{init}} 
	\\ 
	w\mathbf{q}_{1:C}
\end{bmatrix}
\end{equation}
with respect to the template garment vertices $\mat{G}^g$, where the first block $\mathbf{L}^g\mathbf{G}^g = \boldsymbol{\Delta}_{\mathrm{init}}$ forces the solution to keep the local surface structure, while the second block $w\mathbf{I}_{C\times{m_g}}\mat{G}^g = w\mathbf{q}_{1:C}$ makes the boundaries match. 
The nice property of the linear system solve is that the garment template globally stretches or compresses to match the scan garment boundaries, which would take many iterations of non-linear non-rigid registration~\cite{ponsmoll2017clothcap} with the risk of converging to bad local minima. After this initialization, we non-linearly register each garment $\mathbf{G}^g$ to fit the scan surface. We build on top of the proposed multi-part registration in~\cite{ponsmoll2017clothcap} and propose additional loss terms on garment vertices, $v_k \in \mat{G}^g$, to facilitate better garment unposing, $E_{\mathrm{unpose}}$, and minimize interpenetration, $E_{\mathrm{interp}}$, with the underlying SMPL body surface, $\set{S}$.

\begin{equation}
\label{eq:smpl_interepenetration}
    E_\mathrm{{interp}}=\sum_{g}\sum_{\mat{v}_k \in \mat{G}^g} d(\mat{v}_k,\set{S})
\end{equation}
\begin{equation}
d(\mat{x},\set{S})=
\begin{cases}
0, & \text{if }\mat{x}\text{ outside }\set{S} \\
w*|\mat{x}-\mat{y}|_2, & \text{if }\mat{x}\text{ inside }\set{S}
\end{cases}
\end{equation}
where $w$ is a constant ($w = 25$ in our experiments), $\mat{v}_k$ is the $k^{th}$ vertex of $\mat{G}^g$ and $\mat{y}$ is the point closest to $\mat{x}$ on $\set{S}$.

Our garment formulation allows us to freely repose the garment vertices. We can use this to our advantage for applications such as animating clothed virtual avatars, garment re-targeting etc. However, posing is highly non-linear and can lead to undesired artefacts, specially when re-targeting garments across subjects with very different poses. Since we re-target the garments in unposed space, we reduce distortion by forcing distances from garment vertices to the body to be preserved after unposing:
\begin{equation}
    \label{eq:unpose}
    E_{\mathrm{unpose}} = \sum_g \sum_{\mat{v}_k \in \mat{G}^g} (d(\mat{v}_k, \set{S}) - d(\mat{v}^0_k, \set{S}^0))^2
\end{equation}
where $d(\mat{x}, \set{S})$ is the $L_2$ distance between point $\mat{x}$ and surface $\set{S}$. $\mat{v}_k^0$ and $\set{S}^0$ denote garment vertex and body surface in unposed space, using Eq. \ref{eq:garment_pose} and \ref{eq:smplpose} respectively.

\begin{figure*}[t]
	\centering
	\includegraphics[width=\textwidth]{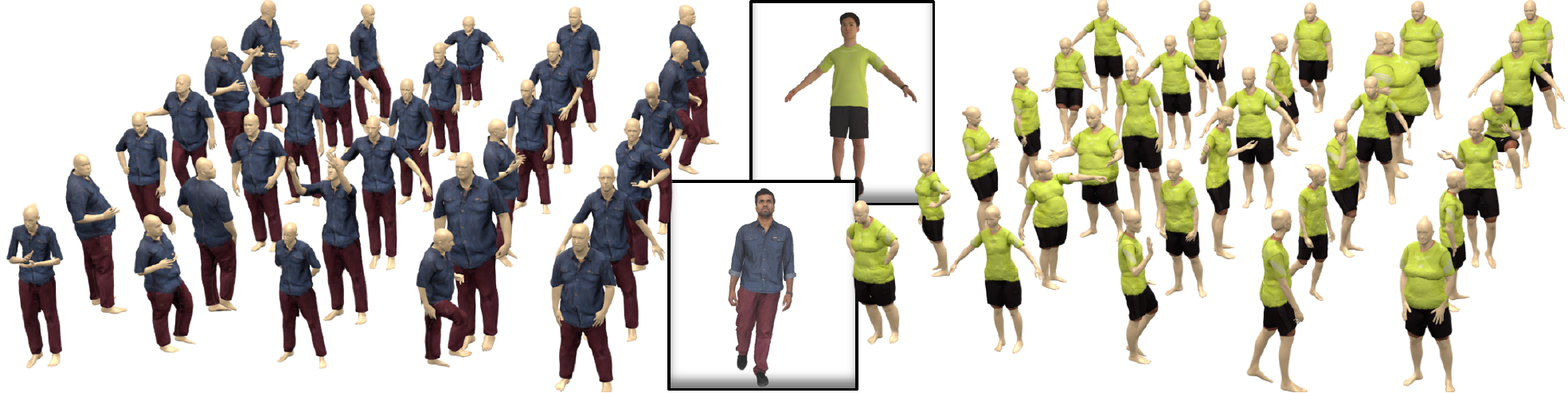}
    \caption{Dressing SMPL with just images. We use MGN to extract garments from the images of a source subject (middle) and use the inferred 3D garments to dress arbitrary human bodies in various poses from SMPL shape subjects. The two sets correspond to male (left) and female (right) body shapes respectively.}
	\label{fig:Dressing_Smpl}
\end{figure*}

\myparagraph{Dressing SMPL}
The SMPL model has proven very useful for modelling unclothed shapes. Our idea is to build a wardrobe of digital clothing compatible with SMPL to model clothed subjects. To this end we propose a simple extension that allows to \emph{dress} SMPL.
Given a garment $\mathbf{G}^g$, we use Eq. \ref{eq:disp}, \ref{eq:garment_shape}, \ref{eq:garment_pose} to pose and skin the garment vertices.
The dressed body including body shape (encoded as $G_1$) will be given by stacking the $L$ individual garment vertices $[G_1(\shape, \pose, \mathbf{D}_1)^T, \hdots ,G_L(\shape, \pose, \mathbf{D}_L)^T]^T$. 
We define the function $C(\pose,\shape,\mat{D})$ which returns the posed and shaped vertices for the skin, and each of the garments combined.\\
See Fig. \ref{fig:Dressing_Smpl} and supplementary for results on re-targeting garments using MGN across different SMPL bodies.

\subsection{From Images to Garments}
\label{sec:MGN}
From registrations, we learn a shape space of garments, and generate a synthetic training dataset with pairs of images and body+3D garment pairs. From this data we train MGN:\emph{Multi-Garment Net}, which maps images to 3D garments and body shape.

\myparagraph{Garment Shape Space} 
In order to factor out pose deformations from garment shape, we ``unpose" the $j^{th}$ garment registrations $\mathbf{G}^g_j\in \mathbb{R}^{m_g\times{3}}$, similar  to ~\cite{shapeundercloth:CVPR17,ponsmoll2017clothcap}. Since the garments of each category are all in correspondence, we can easily compute PCA directly on the unposed vertices to obtain pose-invariant shape basis ($\mat{B}^g$). Using this, we encode a garment shape using $35$ components $\vec{z}^g\in \mathbb{R}^{35}$,
plus a residual vector of offsets $\offsets^{\text{hf},g}_j$,
mathematically: $\mathbf{G}^g_j = \mat{B}^g\mathbf{z}^g_j + \offsets_j^{\text{hf},g}$.
From each scan, we also extract the body shape \emph{under} clothing similarly as in~\cite{shapeundercloth:CVPR17}, which is essential to re-target a garment from one body to another.

\myparagraph{MGN: Multi-Garment Net} The input to the model is a set of semantically segmented images, $\imageset = \{\image_0, \image_1, ..., \image_F-1\}$, and corresponding $2D$ joint estimates, $\jointset = \{\joints_0, \joints_1, ..., \joints_F-1\}$, where $F$ is the number of images used to make the prediction. Following \cite{gong2018instance, alldieck2019learning}, we abstract away the appearance information in RGB images and extract semantic garment segmentation~\cite{gong2018instance} to reduce the risk of over-fitting, albeit at the cost of disregarding useful shading signal. 
For simplicity, let now $\pose$ denote both the joint angles $\pose$ and translation $\trans$. 

The base network, $\basenet$, maps the 2D poses $\jointset$, and image segmentations $\imageset$, to per frame latent code $(\vec{l}_{\poseset})$ corresponding to 3D poses
\begin{equation}
\label{eq:latentcodes_pose}
    \vec{l}_{\poseset} = \basenet^{\theta}(\imageset, \jointset),
\end{equation}
and to a common latent code corresponding to body shape $(\vec{l}_{\beta})$ and garments $(\vec{l}_{\garmparamset})$ by averaging the per frame codes
\begin{equation}
\label{eq:latentcodes_shape_garment}
    \vec{l}_{\shape}, \vec{l}_{\garmparamset} = \frac{1}{F}\sum_{f=0}^{F-1}{\basenet^{\shape, \garmparamset}(\image_f, \joints_f)}.
\end{equation}

For each garment class, we train separate branches, $\garmnet(\cdot)$, to map the latent code $\vec{l}_{\garmparamset}$ to the un-posed garment $\mathbf{G}^g$, which itself is reconstructed from low-frequency PCA coefficients $\mathbf{z}^g$, plus $\mathbf{D}^{\mathrm{hf,g}}$ encoding high-frequency displacements
\begin{equation}
\label{eq:garmentModel}
\garmnet(\vec{l}_{\garmparamset}, \mat{B}^{g}) = \mathbf{G}^g = \mat{B}^g \vec{z}^g +\mathbf{D}^{\mathrm{hf},g}. 
\end{equation}
From the shape and pose latent codes $\vec{l}_{\shape}, \vec{l}_{\pose}$, we predict body shape parameters $\shape$ and pose $\pose$ respectively, using a fully connected layer. Using the predicted body shape $\shape$ and geometry $\garmnet(\vec{l}_{\garmparamset}, \mat{B}^g)$ we compute displacements as in Eq.~\ref{eq:disp}:
\begin{equation}
\label{eq:garmentOffsets}
\offsets^g = \garmnet(\vec{l}_{\garmparamset}, \mat{B}^g) - \mathbf{I}^g\posefun(\shape,\mat{0}_{\mat{\theta}},\mat{0_D}).
\end{equation}
Consequently, the final predicted $3D$ vertices posed for the $f^{th}$ frame are obtained with $C(\shape, \pose_f, \offsets)$, from which we render 2D segmentation masks 
\begin{equation}
\label{eq:rendered}
    \mat{R}_f= R(C(\shape, \pose_f, \offsets), c),
\end{equation}
where $R(\cdot)$ is a differentiable renderer~\cite{henderson18bmvc}, $\mat{R}_f$ the rendered semantic segmentation image for frame $f$, and $c$ denotes the camera parameters that are assumed fixed while the person moves. The rendering layer in Eq.~\eqref{eq:rendered} allows us to compare predictions against the input images.
Since MGN predicts body and garments separately, we can predict a semantic segmentation image, leading to a more fine-grained $2D$ loss, which is not possible using a single mesh surface representation~\cite{alldieck2019learning}. Note that Eq.~\ref{eq:rendered} allows to train with self-supervision. 
\subsection{Loss functions}
\label{sec:losses}
The proposed approach can be trained with $3D$ supervision on vertex coordinates, and with self supervision in the form of $2D$ segmented images. We use upper-hat for variables that are known and used for supervision during training.
We use the following losses to train the network in an end to end fashion:
\begin{itemize}[leftmargin=*]
    \item $3D$ vertex loss in the canonical T-pose ($\pose = \vec{0}_{\pose}$):\\
    \begin{equation}
        \label{eq:TPoseLoss}
        \loss_{\vec{0}_{\pose}}^{3D} = || C(\shape,\vec{0}_{\pose}, \offsets) - C(\hat{\boldsymbol{\beta}},\vec{0}_{\pose}, \hat{\offsets}) ||^2,
    \end{equation}
    where, $\vec{0}_{\pose}$ represents zero-vector corresponding to zero pose.
    \item $3D$ vertex loss in posed space:
    \begin{equation}
        \label{eq:PosedLoss}
    	\small
        \loss_{\poseset}^{3D} = \sum_{f=0}^{F-1} || C(\shape,\pose_f,\offsets) - C(\hat{\shape}, \hat{\pose}_f, \hat{\offsets}) ||^2
    \end{equation}
    \item $2D$ segmentation loss: Unlike \cite{alldieck2019learning} we do not optimize silhouette overlap, instead we jointly optimize the projected per-garment segmentation against the input segmentation mask. This ensures that each garment explains its corresponding mask in the image:
    \begin{equation}
    \label{eq:segmentationLoss}
    \small
    \loss_{seg}^{2D} = \sum_{f=0}^{F-1} || \mat{R}_f - \image_f ||^2,
    \end{equation}
    \vspace{-0.3cm}
    \item Intermediate losses: We further impose losses on intermediate pose, shape and garment parameter predictions:
    $\loss_{\pose} = \sum_{f=0}^{F-1} || \hat{\pose}_f - \pose_f ||^2, \loss_{\shape} = || \hat{\shape} - \shape ||^2,   \loss_{\garmparam} = \sum_{g=0}^{L-1} || \hat{\garmparam}^g - \garmparam^g ||^2$
    where $F, L$ are the number of images and garments respectively. $\hat{\garmparam}$ are the ground truth PCA garment parameters. While such losses are a bit redundant, they stabilize learning.
\end{itemize}

\subsection{Implementation details}
\paragraph{Base Network ($\basenetstar$):} We use a CNN to map the input set $\{\imageset, \jointset\}$ to the body shape, pose and garment latent spaces. It consists of five, $2D$ convolutions followed by max-pooling layers.
Translation invariance, unfortunately, renders CNNs unable to capture the location information of the features. In order to reproduce garment details in 3D, it is important to leverage 2D features as well as their location in the 2D image. To this end, we adopt a strategy similar to \cite{CoordConv2018}, where we append the pixel coordinates to the output of every CNN layer.
We split the last convolutional feature maps into three parts to individuate the body shape, pose and garment information. The three branches are flattened out and we append 2D joint estimates to the pose branch. Three fully connected layers and average pooling on garment and shape latent codes, generate $\vec{l}_{\shape}, \vec{l}_{\pose}$ and $\vec{l}_{\garmparamset}$ respectively. See supplementary for more details.

\myparagraph{Garment Network ($\garmnet$):} We train separate garment networks for each of the garment classes. The garment network consists of two branches. The first predicts the overall mesh shape, and second one adds high frequency details. From the garment latent code ($\vec{l}_{\garmparamset}$), the first branch, consisting of two fully connected layers (sizes=1024, 128), regresses the PCA coefficients. Dot product of these coefficients with the PCA basis generates the base garment mesh. We use the second fully connected branch (size = $m^g$) to regress displacements on top of the mesh predicted in the first branch. We restrict these displacements to $\leq 1cm$ to ensure that overall shape is explained by the PCA mesh and not these displacements.

\section{Dataset and Experiments}
\begin{figure*}[t]
	\centering
	\includegraphics[width=\textwidth]{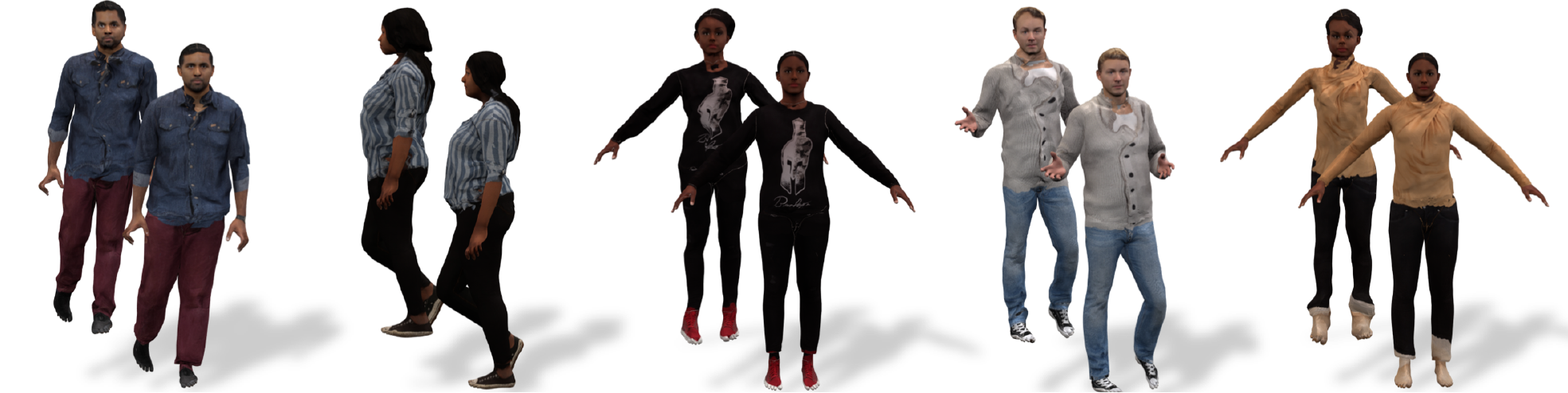}
	\caption{Qualitative comparison with Alldieck \etal \cite{alldieck2019learning}. In each set we visualize 3D predictions from \cite{alldieck2019learning}(left) and our method (right) for five test subjects. Since our approach explicitly models garment geometry, it preserves more garment details, as is evident from minimal distortions across all the subjects. For more results see supplementary. }
	\label{fig:qualitative_scan}
\end{figure*}
\paragraph{Dataset}
We use 356 3D scans of people with various body shapes, poses and in diverse clothing. We held out 70 scans for testing and use the rest for training.
Similar to \cite{alldieck2019learning, alldieck2018video}, we also restrict our setting to the scenario where the person is turning around in front of the camera. We register the scans using multi-mesh registration, SMPL+G. This enables further data augmentation since the registered scans can now be re-posed and re-shaped. \\
We adopt the data pre-processing steps from \cite{alldieck2019learning} including the rendering and segmentation. We also acknowledge the scale ambiguity primarily present between the object size and the distance to the camera. Hence we assume that the subjects in 3D have a fixed height and regress their distance from the camera. Same as \cite{alldieck2019learning}, we also ignore the effect of camera intrinsics.
\begin{figure*}[t]
	\centering
	\includegraphics[width=0.95\textwidth]{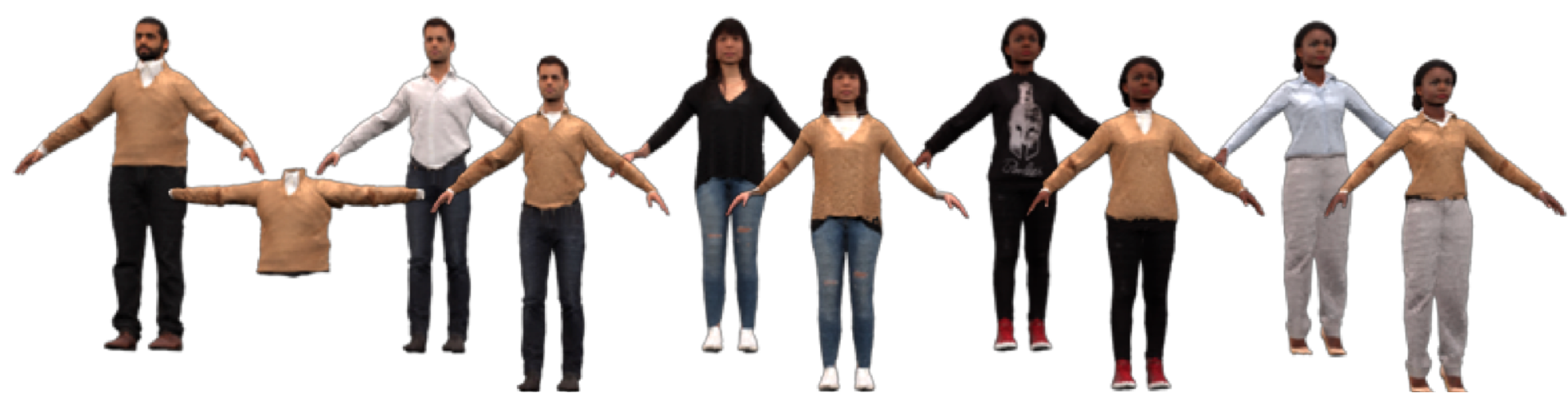}
	\caption{Texture transfer. We model each garment class as a mesh with fixed topology and surface parameterization. This enables us to transfer texture from any garment to any other registered instance of the same class. The first column shows the source garment mesh, while the subsequent images show original and transferred garment texture registrations.}
	\label{fig:tex-transfer}
\end{figure*}
\begin{figure*}[t]
    \centering
    \includegraphics[width=0.95\textwidth]{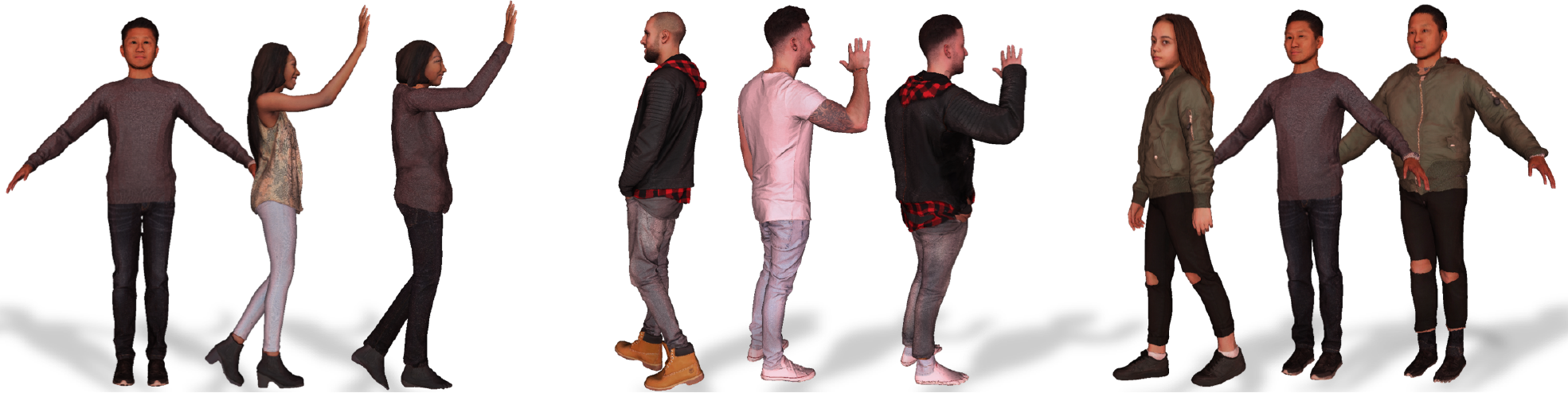}
    \caption{ Garment re-targeting by MGN using 8 RGB images. In each of the three sets we show the source subject, target subject and re-targeted garments. Using MGN, we can re-target garments including both texture and geometry.}
    \label{fig:MGN_retarget}
\end{figure*}
\subsection{Experiments}
In this section we discuss the merits of our approach both qualitatively and quantitatively. We also show real world applications in the form of texture transfer (Fig. \ref{fig:tex-transfer}), where we maintain the original geometry of the source garment but map novel texture. We also show garment re-targeting from images using MGN in Fig. \ref{fig:MGN_retarget}.

\myparagraph{Qualitative comparisons:} We compare our method against \cite{alldieck2019learning} on our scan dataset.
For fair comparison we re-train the models proposed by Alldieck \etal \cite{alldieck2019learning} on our dataset and compare against our approach (Dataset used by \cite{alldieck2019learning} is not publicly available). Figure \ref{fig:qualitative_scan}  indicates the advantage of incorporating the garment model in structured prediction over simply modelling free form displacements. Explicit garment modelling allows us to predict sharper garment boundaries and minimize distortions (see Fig. \ref{fig:qualitative_scan}). More examples are shown in the supplementary material.

\myparagraph{Quantitative Comparison:} In this experiment we do a quantitative analysis of our approach against the state of the art $3D$ prediction method, \cite{alldieck2019learning}. We compute a symmetric error between the predicted and GT garment surfaces similar to \cite{alldieck2019learning}. We report per-garment error, $E^g$ (supplementary), and overall error, i.e. mean of $E^g$ over all the garments
\begin{equation}
\small
    E^g = \frac{1}{N}\sum_{i=1}^N \bigg( \frac{1}{|\mat{\hat{S}}^g_i|}\sum_{\mat{v}_k \in \mat{\hat{S}}^g_i} d(\mat{v}_k, \set{S}^g_i)
    + \\
    \frac{1}{|\mat{S}^g_i|}\sum_{\mat{v}_k \in \mat{S}^g_i} d(\mat{v}_k, \set{\hat{S}}^g_i)
    \bigg),
\end{equation}
where $N$ is the number of meshes with garment $g$. $\mat{S}^g_i$ and $\set{S}^g_i$ denote the set of vertices and the surface of the $i^{th}$ predicted mesh respectively, belonging to garment $g$. Operator $(\hat{.})$ denotes GT values. $d(\mat{v}_k, \set{S})$ computes the $L_2$ distance between the vertex $\mat{v}_k$ and surface $\set{S}$. 

This criterion is slightly different than \cite{alldieck2019learning} because we do not evaluate error on the skin parts.
We reconstruct the 3D garments with mean vertex-to-surface error of 5.78 mm with 8 frames as input. We re-train octopus \cite{alldieck2019learning} on our dataset and the resulting error is 5.72mm.

We acknowledge the slightly better performance of \cite{alldieck2019learning} and attribute it to the fact that the single mesh based approaches do not bind vertices to semantic roles, i.e these approaches can pull vertices from any part of the mesh to explain $3D$ deformations where as our approach ensures that only semantically correct vertices explain the $3D$ shape.

It is also worth noting that MGN predicts garments as linear function (PCA coefficients) of latent code, whereas \cite{alldieck2019learning} deploys GraphCNN. PCA based formulation though easily tractable is inherently biased towards smooth results. Our work paves the way for further exploration into building garment models for modelling the variations in garment geometry over a fixed topology.

We report the results for using varying number of frames in the supplementary.

\myparagraph{GT vs Predicted pose:} The $3D$ vertex predictions are a function of pose and shape. In this experiment we do an ablation study to isolate the effect of  errors in pose estimation on vertex predictions.
This experiment is important to better understand the strengths and weaknesses of the proposed approach in shape estimation by marginalizing over the errors due to pose fitting.
We study two scenarios, first where we predict the 3D pose and second, where we have access to GT pose. We report mean vertex-to-surface error of 5.78mm with GT poses and 11.90mm with our predicted poses.

\subsection{Re-targeting}
Our multi-mesh representation essentially decouples the underlying body and the garments. This opens up an interesting possibility to take garments from source subject and virtually dress a novel subject. Since the source and the target subjects could be in different poses, we first unpose the source body and garments along with the target body. We drop the $(.)^0$ notation for the unposed space in the following section for clarity. Below we propose and compare two garment re-targeting approaches. After re-targeting the target body and re-targeted garments are re-posed to their original poses.

\myparagraph{Naive re-targeting:} The simplest approach to re-target clothes from source to target is to extract the garment offsets, $\offsets^{s,g}$ from the source subject using Eq. \ref{eq:garmentOffsets} and dress a target subject using Eq. \ref{eq:garment_pose}.

\myparagraph{Body aware re-targeting:} The naive approach is problematic because it relies on non-local pre-set vertex association between the garment and the body ($I^g$).
This results in inaccurate association between the body blend shapes, $\offsetfun_{p,s}$ and the garment vertices. This eventually leads to incorrect estimation of source offsets,  $\offsets^{s,g}$ and in turn leads to higher inter-penetrations between the re-targeted garment and the body (see supplementary). In order to mitigate this issue, we compute the new $k^{th}$ target garment vertex location, $\mat{v}_k^t$ as follows
\begin{equation}
    \mat{v}_k^t = \mat{v}_k^s - \mat{S}^{s}_{I_k} + \mat{S}^{t}_{\text{I}_k}
\end{equation}
\begin{equation}
    \text{I}_k = \underset{\text{I} \in [0,\text{ }|\mat{S}^{s}|-1] }{\text{argmin}} \text{ }||\mat{v}_k^s - \mat{S}^s_{\text{I}}||_2,
\end{equation}
where $\mat{v}_k^s$ is the source garment vertex, $\mat{S}^s_{\text{I}_k}$ is the vertex (indexed by $\text{I}_k$) among the source body vertices, $\mat{S}^{s}$, closest to $\mat{v}_k^s$ and $\mat{S}^{t}_{\text{I}_k}$ is the corresponding vertex among the target body vertices.

MGN allows us to predict separable body shape and garments in 3D, allowing us to do garment re-targeting (as described above) using just images. To the best of our knowledge this is the first method to do so. See Fig. \ref{fig:MGN_retarget} for results on garment re-targeting by MGN. See supplementary for more results.

\section{Conclusion and Future Works}
We introduce MGN, the first model capable of jointly reconstructing from few images, body shape and garment geometry as layered meshes. 
Experiments demonstrate that this representation has several benefits: it is closer to how clothing layers on top of the body in the real world, which allows control such as re-dressing novel shapes with the reconstructed clothing. Additionally, we introduce for the first time, a dataset of registered \emph{real} garments from real scans obtained with a robust registration pipeline.
When compared to more classical single mesh representations, it allows more control and qualitatively the results are very similar. 
In summary, we think that MGN provides a first step in a promising research direction. We will release the MGN model and the digital wardrobe to stimulate research in this direction.
Further discussion on limitations and future works in supplementary.

\myparagraph{Acknowledgements} This work is partly funded by the Deutsche Forschungsgemeinschaft (DFG, German Research Foundation) - 409792180 (Emmy Noether Programme, project: Real Virtual Humans) and Google Faculty Research Award. 
We thank twindom (https://web.twindom.com/) for providing scan data, Thiemo Alldieck for providing code for texture/segmentation stitching, and Verica Lazova for discussions.

{\small
\bibliographystyle{ieee_fullname}
\bibliography{egbib}
}

\end{document}